\documentclass[19pt,conference]{IEEEtran}
\usepackage{amsmath, amsfonts, amssymb, amsthm} %
\usepackage{array}
\usepackage{graphicx}
\usepackage{xspace}
\usepackage[sort&compress, numbers]{natbib}
\usepackage{stmaryrd}
\usepackage{xcolor}
\usepackage{mathtools}
\usepackage{float}
\usepackage{textcomp}
\usepackage{multirow}
\usepackage{diagbox}
\usepackage[printonlyused,withpage,nolist,nohyperlinks]{acronym}
\usepackage{subcaption}
\usepackage{caption}
\usepackage{lipsum}
\usepackage{algorithm}
\usepackage{algpseudocode}  
\usepackage{stfloats}
\usepackage{placeins}
\usepackage{url}
\usepackage{microtype}

\begin{document}
\title{Context-Aware Behavior Learning with Heuristic Motion Memory for Underwater Manipulation}
\author{\IEEEauthorblockN{ 
        Markus Buchholz\IEEEauthorrefmark{1}, 
        Ignacio Carlucho\IEEEauthorrefmark{1},
        Michele Grimaldi\IEEEauthorrefmark{1},
        Maria Koskinopoulou\IEEEauthorrefmark{1} and
        Yvan R. Petillot\IEEEauthorrefmark{1}
    }
    \IEEEauthorblockA{
        \IEEEauthorrefmark{1}School of Engineering \& Physical Sciences, Heriot-Watt University, Edinburgh, UK\\
       }
}
\maketitle
\begin{abstract}
Autonomous motion planning is critical for efficient and safe underwater manipulation in dynamic marine environments. 
Current motion planning methods often fail to effectively utilize prior motion experiences and adapt to real-time uncertainties inherent in underwater settings. 
In this paper, we introduce an Adaptive Heuristic Motion Planner 
framework that integrates a Heuristic Motion Space (HMS) with Bayesian Networks to enhance motion planning for autonomous underwater manipulation. %
Our approach employs the Probabilistic Roadmap (PRM) algorithm within HMS to optimize paths by minimizing a composite cost function that accounts for distance, uncertainty, energy consumption, and execution time. By leveraging HMS, our framework significantly reduces the search space, thereby boosting computational performance and enabling real-time planning capabilities. 
Bayesian Networks are utilized to dynamically update uncertainty estimates based on real-time sensor data and environmental conditions, thereby refining the joint probability of path success. 
Through extensive simulations and real-world test scenarios, we showcase the advantages of our method in terms of enhanced performance and robustness. This probabilistic approach significantly advances the capability of autonomous underwater robots, ensuring optimized motion planning in the face of dynamic marine challenges.
\end{abstract}

\FloatBarrier
\IEEEtriggeratref{1}

\section{Introduction}

Effective motion planning in challenging underwater environments is critical, especially in scenarios like underwater construction or inspection, where nonlinear dynamics, unpredictable disturbances, and sensor noise complicate navigation. For instance, underwater welding \cite{weld_1, weld_2} requires precise and often repeated manipulator motions over extended operational periods in dynamic conditions. Traditional motion planning techniques—such as inverse kinematics and optimization-based methods—are typically executed offline, which can hinder real-time responsiveness when unexpected changes occur \cite{art_22,art_23,art_24}. Furthermore, repeatedly re-planning complex motion trajectories in such tasks using conventional methods can become computationally expensive and time-consuming.

\begin{figure}[htbp] 
    \centering
    \includegraphics[width=0.45\textwidth]{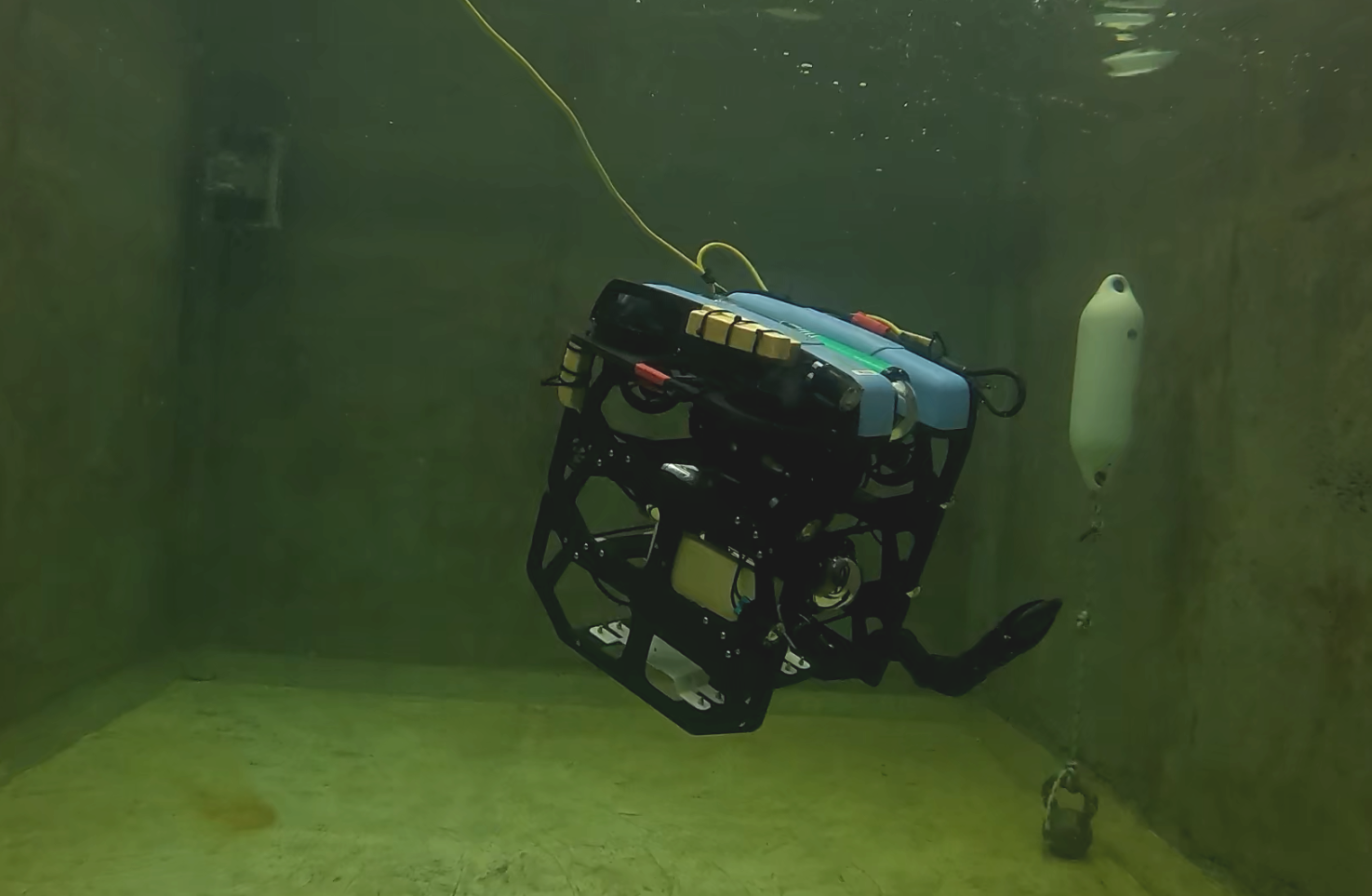}
    \caption{Experimental setup of the UVMS used during the experiments at our Heriot-Watt University Laboratory.}
    \label{fig:intro}
\end{figure}

Underwater Vehicle Manipulator System (UVMS) equipped with robotic manipulators (Fig.~\ref{fig:intro}) have become essential assets for a variety of offshore and deep-sea applications, including inspection, maintenance, and complex manipulation tasks \cite{art_1,art_20,morgan2022autonomous}. 
These operations demand a high level of autonomy and efficiency, particularly in dynamic and unstructured underwater environments where conditions are highly unpredictable \cite{art_21}.
Furthermore, many AUV deployments involve tethered configurations, which introduce additional constraints on motion and maneuverability, requiring robust and adaptable planning algorithms \cite{yoerger1990design, whitcomb2000design, allotta2011tethered}.
Recent advancements have sought to address the limitations of traditional motion planning techniques by leveraging prior motion experiences and probabilistic reasoning to enhance autonomous decision-making. However, existing methods often struggle to balance computational efficiency with the ability to adapt to real-time environmental changes. For instance, Experience-Based Bidirectional RRT (EB-RRT) \cite{art_9} utilizes stored experience graphs to accelerate path planning in semi-structured environments but lacks the capability for real-time adaptability and robust uncertainty handling. Similarly, learning-based approaches such as N2M2 \cite{art_11} and HyperPlan \cite{art_10} offer adaptability through reinforcement learning and hyperparameter optimization; meanwhile, trajectory optimization methods like CHOMP \cite{chomp} and sampling-based techniques such as SANDROS \cite{sandros} have been developed to refine path smoothness and address dynamic uncertainties. However, these approaches are often resource-intensive and less suited for real-time applications in dynamic settings.

In contrast, sampling-based methods such as the Probabilistic Roadmap (PRM)\cite{prm} algorithm have become widely adopted for their ability to rapidly construct a collision-free graph of the robot’s configuration space by randomly sampling potential configurations. 
However, in cluttered environments—where obstacles create narrow passages and complex connectivity patterns—a higher number of samples is generally required to ensure that the roadmap is dense enough to capture the intricacies of the configuration space. This increase also enlarges the search space, leading to higher computational overhead. Traditional PRM combined with pure A* search therefore faces a trade-off: while increasing \textit{max\_samples} improves coverage and connectivity, it also tends to degrade real-time performance due to the expanded number of nodes that must be processed.

To address this challenge, we propose an Adaptive Heuristic Motion Planner (AHMP), an enhanced framework that augments the conventional PRM with Heuristic Motion Space (HMS). Our approach retains the extensive sampling benefits of PRM—ensuring robust coverage even when a high \textit{max\_samples} is required—while mitigating the increased computational effort typically imposed on A* search. HMS achieves this by caching and reusing previously computed motion plans, effectively creating a \emph{memory} of high-value nodes. %
These cached experiences allow the planner to bypass large portions of the roadmap, thereby reducing the effective search space and yielding more stable and efficient performance in real-time.
Furthermore, our integration of Bayesian Networks within HMS dynamically refines uncertainty estimates based on real-time sensor data and environmental conditions. This probabilistic reasoning is not solely focused on minimizing distance or cost but also on maximizing the likelihood of a path's success. In this way, the system adapts to the inherent uncertainty of underwater environments by favoring pathways that are both short and reliably safe.

The key contributions of this research are:
\begin{enumerate}
    \item The development of an HMS framework that leverages stored motion experiences to significantly reduce the effective search space in PRM.
    \item The integration of Bayesian Networks for dynamic uncertainty management, enabling probabilistic assessments that guide path selection.
    \item A novel hybrid approach that maintains robust roadmap connectivity—achieved by high \textit{max\_samples}—while ensuring real-time performance through memory-based reuse of previously computed paths.
\end{enumerate}

\section{Related Work}

Experience-based motion planning methods have demonstrated significant potential for accelerating repetitive tasks by reusing past solutions. For instance, Experience-based Bidirectional RRT (EB-RRT) \cite{art_9} employs stored experience graphs and adaptive sampling techniques to enhance computational efficiency in semi-structured environments. However, its reliance on static experience graphs limits its adaptability to dynamic and uncertain scenarios. Similarly, Motion Memory \cite{art_5} leverages past trajectories to bias sampling and accelerate motion planning in similar environments but struggles to generalize across different conditions or handle real-time uncertainties. While these frameworks successfully utilize past motion experiences, they do not integrate mechanisms for probabilistic reasoning or dynamic adaptation.

Additionally, the Experience-based Subproblem Planning (E-ARC) approach \cite{art_27} tackles multi-robot motion planning by leveraging a database of precomputed solutions for lower-dimensional subproblems. E-ARC efficiently resolves local conflicts by retrieving relevant solutions from the database, significantly improving scalability in multi-robot scenarios. However, its reliance on subproblem decomposition makes it less suitable for environments with high-dimensional continuous planning requirements, such as those encountered in underwater manipulation tasks. Moreover, it does not incorporate probabilistic reasoning for handling dynamic uncertainties.

Another innovative method, the 3D-CNN-Based Heuristic Guided Task-Space Planner (HM-TS-RRT) \cite{art_28}, integrates task-space planning with heuristic maps generated through deep learning. This framework excels in leveraging environmental information to guide exploration and exploitation, reducing planning time and improving success rates in complex environments. However, its dependence on pre-trained heuristic maps limits its adaptability to unstructured and dynamically changing environments, as it lacks mechanisms for real-time uncertainty updates.

Probabilistic frameworks, such as the Dynamic Bayesian Threat Assessment Framework \cite{art_6} and the Bayes Adaptive MDP Model \cite{art_13}, have been effective in addressing uncertainties in motion planning. The former focuses on reactive threat assessment for unmanned underwater vehicles, while the latter emphasizes state transition modeling to optimize long-term decisions. However, these methods often lack mechanisms to leverage prior motion experiences, leading to increased computational costs for real-time operations. HMS bridges this gap by combining heuristic motion reuse with Bayesian reasoning, enabling both proactive and reactive planning to ensure safety and efficiency.

Learning-based methods, have advanced the adaptability of motion planning through reinforcement learning and algorithmic optimization \cite{carlucho2020reinforcement}. N2M2 \cite{art_11}  leverages reinforcement learning to navigate unseen environments dynamically, but its reliance on extensive training datasets makes it resource-intensive and less suitable for resource-constrained underwater systems. Similarly, HyperPlan’s hyperparameter optimization approach is tailored for static planning problems \cite{art_10}, limiting its applicability in dynamic and uncertain scenarios. By avoiding the overhead of training while continuously updating motion heuristics in real time, HMS provides a lightweight and efficient alternative for underwater operations.

Recent research on multi-heuristic planning, exemplified by Multi-Heuristic A* \cite{art_4} and MR-MHA* \cite{art_18}, has shown promise in high-dimensional spaces by combining diverse heuristics for pathfinding. While these methods improve computational efficiency, they do not incorporate memory-based reasoning or uncertainty updates, which are integral to the HMS framework. Furthermore, task-space planners such as the Task-Space Motion Primitive Framework \cite{art_17} focus on geometric reasoning to navigate cluttered environments but lack the adaptability provided by probabilistic reasoning and dynamic memory updates.

Other novel approaches, such as vectorized sampling-based methods (e.g., Motions in Microseconds \cite{art_18}) and semantic knowledge-based planning \cite{art_12}, focus on computational speed and contextual understanding, respectively. While these methods excel in specific applications, they are limited in their ability to adapt to dynamic environments and incorporate prior experiences.

Unlike existing approaches that rely on static assumptions or precomputed maps, our proposed AHMP enables the transfer of heuristic motion spaces to unexplored environments through dynamic adaptation to environmental uncertainties. This integration of heuristic motion reuse with probabilistic updates creates a comprehensive framework that simultaneously achieves efficiency, adaptability, and safety, making it particularly effective for underwater robotic manipulation in dynamic and uncertain environments.

\begin{figure*}[t]
\centering
\includegraphics[width=0.85\textwidth, trim={1.5cm 6.5cm 2cm 0cm},clip]{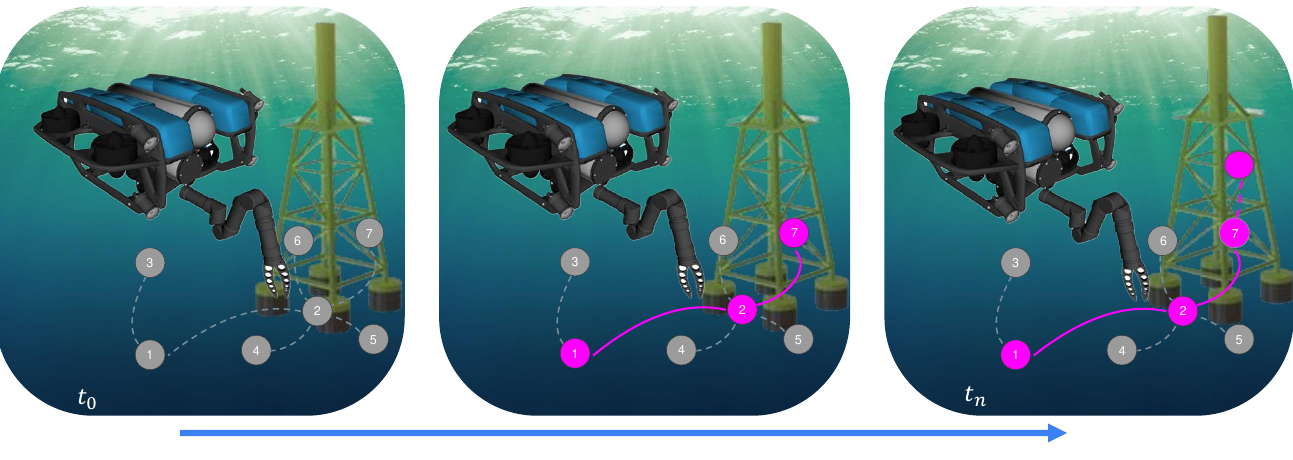}
\caption{Conceptual overview of the AHMP. The algorithm starts by building a PRM (Left). As more paths are explored, a memory of high value nodes is created which allows to reduce the search space. When new goals are introduced, the previous experience can be used to achieve faster and safer motion plans.} 
\label{fig:idea}
\end{figure*}

\section{Methodology}
\label{sec:proposed_algorithm}
This section provides a description of the proposed methodology, the AHMP, which efficiently plans paths by building and reusing motion experiences. A high-level overview of the AHMP concept is illustrated in Fig. \ref{fig:idea}, followed by a detailed description of the algorithmic procedure.

\subsection{Overview}

Motion planning with PRM is typically conducted in two stages. First, a PRM is constructed by randomly sampling up to \textit{max\_samples} collision-free configurations and connecting them via feasible local paths. Then a search method, like A*, is used to find a suitable path. On difficult spaces or with high density of objects, a high \textit{max\_samples} is used to ensure that narrow passages and intricate maneuvers can be captured, but it also increases the number of nodes that a graph-based search (e.g., A*) must explore.

Rather than reducing the local density of PRM, we propose to create a hierarchy where we layer a HMS on top of PRM to store what we refer to as \emph{highway nodes}. 
These nodes act as efficient shortcuts that complement the full, densely sampled PRM. Because the base roadmap is constructed with a high \textit{max\_samples}, each highway path inherits the fine-grained resolution around obstacles, thus maintaining safe clearance. By focusing the A*-based query on these highways, the search expands fewer nodes overall, which is crucial in to achieve fast and safe planning in large-scale or cluttered underwater environments.

The overall proposed framework is presented in Figure \ref{fig:arch}. The Adaptive Heuristic Motion Planner (AHMP) operates at two hierarchical levels: a High-Level Planner (HMS + BN) layered on top of a low-level PRM. First, a PRM is constructed in the manipulator's configuration space (or the combined vehicle-manipulator space). This involves sampling collision-free configurations and connecting them with feasible local paths to form a graph.  Then, a HMS repository, denoted as \(\mathcal{H}\), is built to store selected paths and frequently used sub-paths as motion primitives \(\{M_i\}_{i=1}^n\). Each motion primitive \(M_i\) in HMS is associated with an uncertainty estimate \(U_i\) and other relevant execution metrics.

\subsection{System Architecture}

\begin{figure}[t]
\centering
\includegraphics[width=0.5\textwidth]{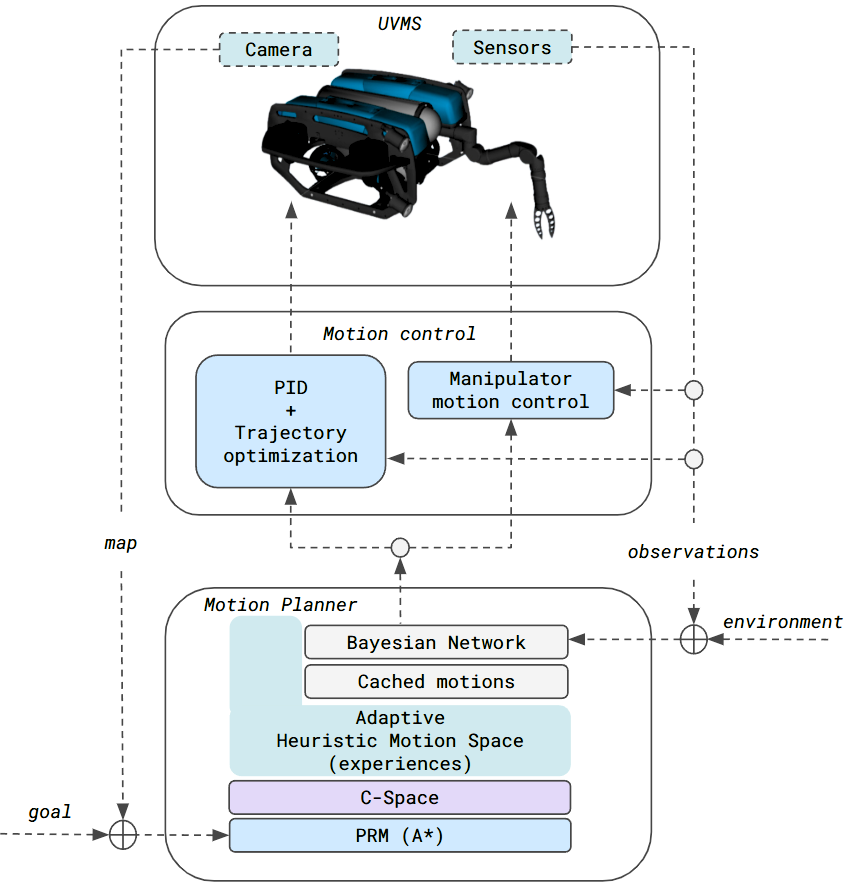}
\caption{The general architecture of a UVMS outlining the interaction among the manipulator (Reach Alpha 5), the underwater vehicle (BlueROV2), and the integrated motion control and planning. }
\label{fig:arch}
\end{figure}

A comprehensive system architecture, illustrating the integrated components and data flow, is depicted in Fig.~\ref{fig:arch}. The system is designed for the BlueROV2 underwater vehicle platform \cite{bluerobotics} and the Reach Alpha 5 manipulator \cite{reach}, forming the UVMS.

The architecture comprises three primary modules: the UVMS platform, the Motion Control module, and the Motion Planner module. The UVMS platform encapsulates the physical vehicle and its sensors, including a \textit{Camera} for visual input and a suite of \textit{Sensors} (DVL, IMU, sonar, etc.) for environmental awareness.

The Motion Control module receives trajectory commands from the Motion Planner and translates them into actuation signals. It consists of a cascaded PID controller for vehicle positioning and a dedicated manipulator motion controller, driven by an Inverse Kinematics (IK) solver, for precise joint angle management. The \textit{observations} from the UVMS sensors, including visual and environmental data, are fed back to the Motion Control module to enable closed-loop control and trajectory optimization. The Motion Control module also receives an \emph{external} signal, representing potential external disturbances or commands.

The Motion Planner module is the core of the proposed Adaptive Heuristic Motion Planner (AHMP). It incorporates a C-Space representation, a PRM with A* search for path planning, and a HMS for storing and retrieving motion experiences. The HMS includes \textit{Cached motions} and a \textit{BN}. The \textit{observations} from the UVMS sensors are used to update the BN, which in turn monitors system uncertainties and environmental conditions, allowing the AHMP to dynamically adapt its path planning. The environmental \emph{map}, derived from sensor data, is used to update the HMS and refine the uncertainty estimates within the BN. The \emph{goal} configuration provides the target for the Motion Planner. The AHMP generates optimized trajectories which are passed to the Motion Control module.

The interconnected nature of these modules, facilitated by the exchange of signals such as \textit{observations}, \emph{map}, \emph{goal}, and \emph{external}, enables a closed-loop feedback system. This system empowers the UVMS to learn from experience, adaptively improve its motion primitives, and enhance both efficiency and reliability in underwater navigation and manipulation tasks.

\begin{algorithm}[t]
\caption{Adaptive Heuristic Motion Planner (AHMP)}
\label{alg:path_planning}
\small
\begin{algorithmic}[1]
    \Require
    $\mathcal{E},\,s,\,\{g_i\} = \{g_1, \dots, g_n\},\,\mathrm{HMS},\,\mathrm{BN},\,\tau$
    \Ensure
    \(\{\Pi_i\}\): Paths from \(s\) to each \(g_i\)

    \State Initialize paths \(\Pi_i \gets \varnothing\) for all \(g_i\)
    \State \(\textit{currentNode} \gets s\)

    \For{each \(g_i\) in \(\{g_i\}\)}
        \State Update \(\mathrm{BN}\) with latest sensor data
        \State \(\textit{oNode} \gets \arg\max\limits_{u \in \mathrm{HMS}} \Bigl\{ P(u \mid g_i, \mathrm{BN})\Bigr.\) \(\Bigl. : \text{heuristic}(u, g_i)\leq\tau \Bigr\}\)
        \If{\(\textit{oNode} \neq -1\)}
            \State \(\Pi_{\mathrm{HMS}} \gets \mathrm{HMS}[\textit{oNode}].\text{path}\)
            \State \(\textit{PartialA}\gets \mathrm{A^*}(\mathcal{E}, \textit{currentNode}, \textit{oNode})\)
            \State \(\textit{PartialB}\gets \mathrm{A^*}(\mathcal{E}, \textit{oNode}, g_i)\)
            \If{\(\textit{PartialA}.\text{path} \neq \varnothing \,\wedge\, \textit{PartialB}.\text{path} \neq \varnothing\)}
                \State \(\Pi_i \gets \textit{PartialA}.\text{path} \,\Vert\, \Pi_{\mathrm{HMS}} \,\Vert\, \textit{PartialB}.\text{path}\)
            \Else
                \State \(\textit{FullResult} \gets \mathrm{A^*}(\mathcal{E}, \textit{currentNode}, g_i)\)
                \State \(\Pi_i \gets \textit{FullResult}.\text{path}\)
            \EndIf
        \Else
            \State \(\textit{FullResult} \gets \mathrm{A^*}(\mathcal{E}, \textit{currentNode}, g_i)\)
            \State \(\Pi_i \gets \textit{FullResult}.\text{path}\)
        \EndIf

        \If{\(\Pi_i \neq \varnothing\)}
            \State \(\mathrm{HMS}[g_i] \gets (\Pi_i, 1.0)\)  \Comment{Cache new path}
            \For{each \(u\) in \(\mathrm{HMS}\)}
                \State \(p_{\mathrm{new}}(u) \;\propto\; p_{\mathrm{old}}(u)\times\exp\bigl(-\alpha\|\Pi_i\|\bigr)\)
            \EndFor
            \State Normalize \(p_{\mathrm{new}}(u)\)
            \State Update \(\mathrm{BN}\) with new path data
            \State \(\textit{currentNode} \gets g_i\)
        \EndIf
    \EndFor

    \Return \(\{\Pi_i\}\)
\end{algorithmic}
\end{algorithm}

\subsection{Adaptive Heuristic Motion Planner}
 
The key objective of our method is to generate a safe path from the current state, C, to the end-effector goal configuration, \(G = (x_g, y_g, z_g)\).  

Our method starts by constructing a PRM by randomly sampling up to \textit{max\_samples} collision-free configurations and connecting them via feasible local paths. As previously stated, when the PRM is dense and queries are frequent, naive A* can become computationally expensive due to its exponential complexity.
To mitigate this issue, we layer the HMS on top of PRM to store \emph{highway nodes}, key configurations frequently used in successful paths, and \emph{cached paths} connecting them. These highway nodes act as efficient shortcuts that complement the full, densely sampled PRM.

When tasked with reaching a goal configuration \(G = (x_g, y_g, z_g)\), AHMP will first utilize PRM plus A* to find an optimal trajectory. This expansion is only done when necessary, for example, on first expansions,  thereby minimizing computational overhead. 
After generating initial trajectories, AHMP efficiently reuses precomputed paths stored in the HMS. A BN integrates data related to external environmental changes to probabilistically select the most optimal paths, ensuring both enhanced performance and robust adaptation.
The evaluation of candidate HMS paths is based on the current environmental state \(E\).  Specifically, for each stored motion primitive \(M_i\), the probability of it leading to the goal \(G\) given the environment \(E\) is calculated as:
\begin{equation*}
\begin{split}
P(M_i \mid G, E) \;=\;& \frac{P\bigl(G \mid M_i, E\bigr)\,P\bigl(M_i \mid E\bigr)}{P(G \mid E)} \\
& \propto \;\frac{\exp(-\lambda\,U_i)}{\,1 + d(M_i, G)\,},
\end{split}
\end{equation*}
where \(d(M_i, G)\) is a distance metric in configuration space (joint or Euclidean space), and \(\lambda\) is a confidence parameter.  A threshold \(\tau\) is used to define the maximum allowable distance from the goal within which HMS nodes are considered as potential approach points.

The BN used in our work is a Directed Acyclic Graph (DAG). The BN nodes \(X_1,\dots,X_n\) represent random variables, and its edges encode conditional dependencies. The joint distribution can be factorized as:
\[
P(X_1,\dots,X_n)
\;=\;
\prod_{i=1}^{n}
  P\bigl(X_i \mid \mathrm{Pa}(X_i)\bigr),
\]
where \(\mathrm{Pa}(X_i)\) denotes the parents of \(X_i\). By integrating the BN with HMS and the underlying PRM, the algorithm dynamically identifies and prioritizes those HMS nodes most likely to yield an optimal path under current conditions. This BN-based evaluation enhances both computational efficiency (by pruning less-promising expansions) and trajectory reliability (by favoring paths robust to disturbances).

The algorithm's step-by-step procedure is detailed in the pseudocode provided in Algorithm \ref{alg:path_planning}.  The algorithm takes as input the environment (\(\mathcal{E}\)), the starting state (\(s\)), a list of goal configurations (\(\{g_i\}\)), the HMS, BN, and a distance threshold (\(\tau\)). It outputs a list of paths (\(\{\Pi_i\}\)) from the start to each goal.

In operation, for each goal, the algorithm first updates the BN with the latest sensor data. It then identifies the optimal approach node (\(\textit{oNode}\)) within the HMS by selecting the node that maximizes the probability of reaching the goal, given the BN and ensuring it is within the distance threshold \(\tau\) from the goal.
 
It then performs A* search in two segments: from the current configuration \(C\) to \(\textit{oNode}\), and from \(\textit{oNode}\) to \(g_i\). Should either of these partial searches fail, the algorithm reverts to a full A* search directly from \(C\) to \(g_i\) on the PRM. 
Following successful trajectory generation, the motion experience is cached in the HMS to facilitate future planning. The BN then updates the probabilities of these cached motions, incorporating the environmental impact and refining the selection of experiences most likely to yield optimal paths.
This entire process ensures that the system adaptively leverages historical motion data to accelerate path planning while maintaining robustness through the underlying PRM and probabilistic evaluation.

\subsection{Computation calculation}

Utilizing PRM with a naive A* can become computationally expensive due to its exponential complexity. The complexity in this case can be measured as 
\(O(b^d)\), where \(b\) is the branching factor and \(d\) is the effective search depth. If an agent needs to reach \(N\) distinct goals, naive repeated A* on the PRM leads to a total time of \(N \cdot O\bigl(b^d\bigr)\). By contrast, our proposed method has a complexity of:
\[
N \cdot O(\lvert \mathrm{HMS} \rvert) \;+\; \sum_{i=1}^N O\bigl(b^{d'_i}\bigr),
\]
where \(\lvert \mathrm{HMS} \rvert\) is the size of the precomputed set of highways and \(d'_i \ll d\) is the reduced effective depth. The ratio of times can be approximated as:
\[
\frac{T_{\text{A*}}}{T_{\mathrm{HMS}}}
\;\approx\; \frac{b^d}{\,b^{d'}}
\]
Hence, reusing cached paths in repeated or complex queries substantially mitigates the exponential blow-up otherwise encountered by naive A*.

\section{Experiments}
To validate the algorithm, we run experiments in a tank using a BlueROV2 equipped with a Reach Alpha 5 manipulator as presented in Fig. \ref{fig:intro}. Further experiments were conducted in simulation for comparative evaluation.

The algorithms evaluated in this section were implemented in C++ with a kinematic solver integrated into the open-source Ceres Solver \cite{Ceres} library—and executed on a Linux Ubuntu~24.04 system equipped with an Intel(R) Core(TM) i9-14900K processor and 128\,GB of RAM.

\subsection{Comparative evaluation}
\label{sec:performance-analysis}

We evaluated the performance of our proposed HMS algorithm, alongside baseline A* and RRT approaches, within the UVMS simulation environment depicted in Fig.~\ref{fig:arch}. This environment featured two red obstacles, creating a realistic scenario for assessing path planning capabilities. The performance was analyzed under varying values of the \textit{max\_samples} parameter, which controls the number of collision-free nodes sampled via the PRM method, and \textit{max\_iter\_rrt} for RRT, which dictates the number of iterations to grow the exploration tree. These parameters directly impact the roadmap's coverage and the computational effort required for each path query. Our kinematic model addressed a 5DOF problem, comprising a 4DOF robot arm mounted on a 1DOF floating base. Each test was averaged over 5 runs to ensure statistical robustness.

\begin{figure}[t]
    \centering
    \includegraphics[width=\linewidth]{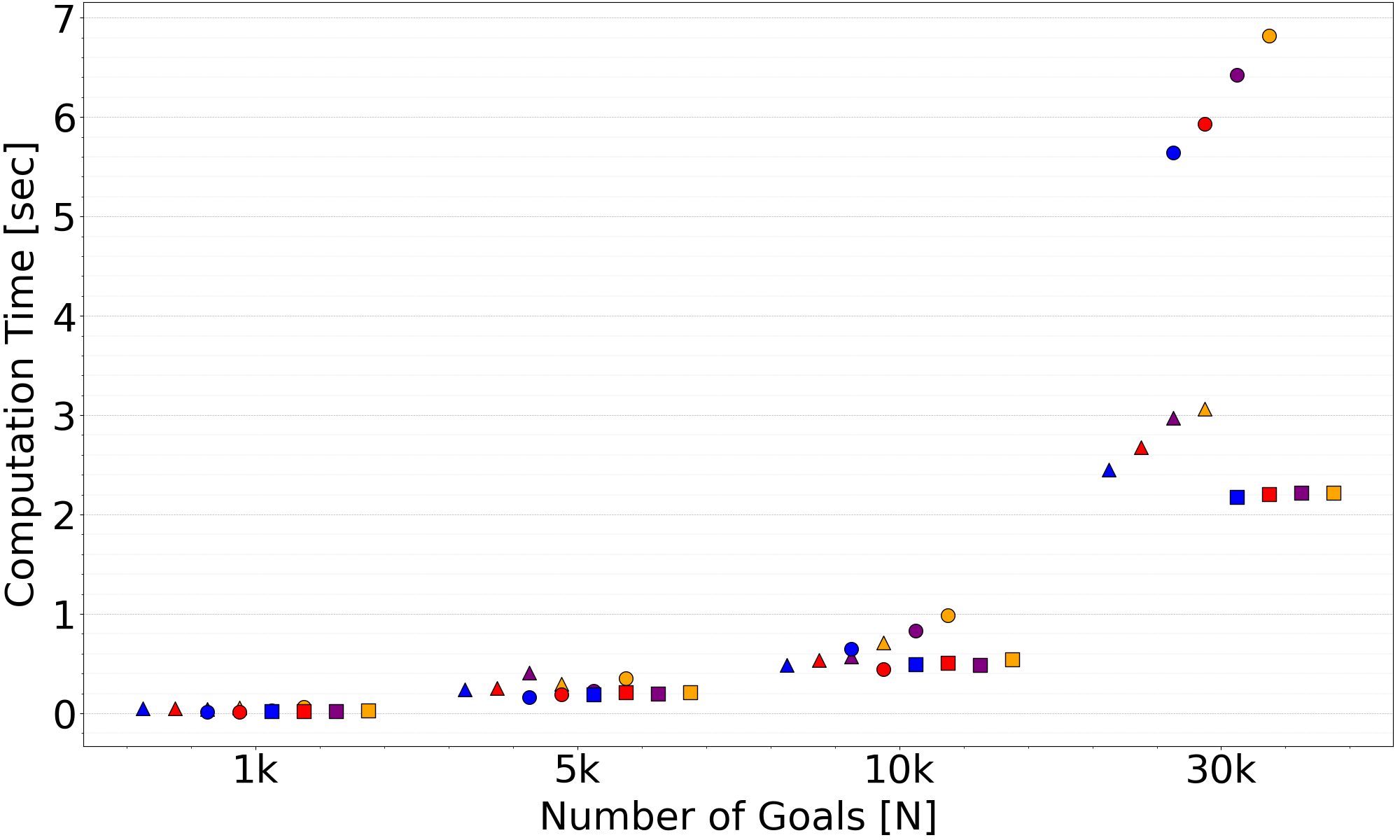}
    \caption{Comparison of execution times for pure PRM\,+\,A* (marked by `$\circ$'), RRT (marked by `$\triangle$'), AHMP (marked by `$\square$') across varying numbers of goals (horizontal axis). Colors indicate different \textit{max\_samples} values: blue for 1{,}000, red for 5{,}000, violet for 10{,}000, and orange for 30{,}000.}
    \label{fig:performance_comparison}
\end{figure}

Results presented in Fig.~\ref{fig:performance_comparison} show that when \textit{max\_samples} is relatively low (e.g., 1{,}000) and the number of goals remains small, pure PRM+A* can be slightly faster, as AHMP introduces overhead from maintaining a hierarchical \emph{memory} and updating Bayesian probabilities. However, as \textit{max\_samples} grows (e.g., to 5{,}000, 10{,}000, or even 30{,}000) and the number of goals increases, AHMP demonstrates clear advantages.
Furthermore, RRT's performance varied. At lower sample/iteration counts, it was sometimes faster, reflecting its ability to quickly find feasible solutions. However, at higher counts, both PRM-based methods (AHMP and A*) generally outperformed RRT, demonstrating superior solution quality and consistency. Notably, AHMP exhibited the most stable performance with increasing goal counts, showcasing the benefits of its cached path mechanism. 

With reward to runtimes, pure PRM + A* scales almost linearly with increasing \textit{max\_samples} and goal counts—rising from about 0.1613\,s at 5{,}000 samples to 5.6457\,s at 30{,}000 samples. On the other hand, AHMP exhibits significantly more stable performance, culminating in only 2.2045\,s at 30{,}000 samples. This trend highlights the efficacy of the hierarchical caching mechanism in reducing the cost of searching through large, densely sampled roadmaps. 
This can be critical in cluttered environments, where higher \textit{max\_samples} values are crucial for discovering feasible paths. These findings reaffirm that while A* may be preferable for minimal sampling and few goals, the AHMP approach offers superior scalability and computational efficiency for navigating large-scale roadmaps with extensive goal sets.
Furthermore, for robot manipulators operating in cluttered environments, such as our simulation, PRM-based methods are generally recommended over RRT. PRM builds a global roadmap that systematically samples the free configuration space, which is particularly advantageous when obstacles create narrow passages and complex connectivity that must be captured accurately.

\subsection{Experimental Results}

The primary objectives of these experiments were to validate the effectiveness of the AHMP in a realistic underwater environment and to quantitatively compare its performance against established path planning algorithms, namely PRM and RRT. We aimed to demonstrate that the AHMP, by leveraging cached motion experiences and adaptive learning, could achieve comparable accuracy to PRM while offering significant advantages in terms of computational efficiency and robustness.

In these experiments, we evaluated the performance of the integrated AHMP approach for planning collision-free trajectories and compared it against both a baseline PRM and the RRT algorithm.

Experiments were conducted in a tank environment (3.5\,m x 3.0\,m x 2.5\,m) that provides a repeatable setup for testing.

In all tests, the vehicle operates in a stabilized ROV mode, ensuring that the motion of the manipulator is decoupled from the vehicle’s movement.

\begin{figure*}[tp]
    \centering
    \begin{subfigure}[t]{0.18\textwidth}
        \centering
        \includegraphics[width=\textwidth]{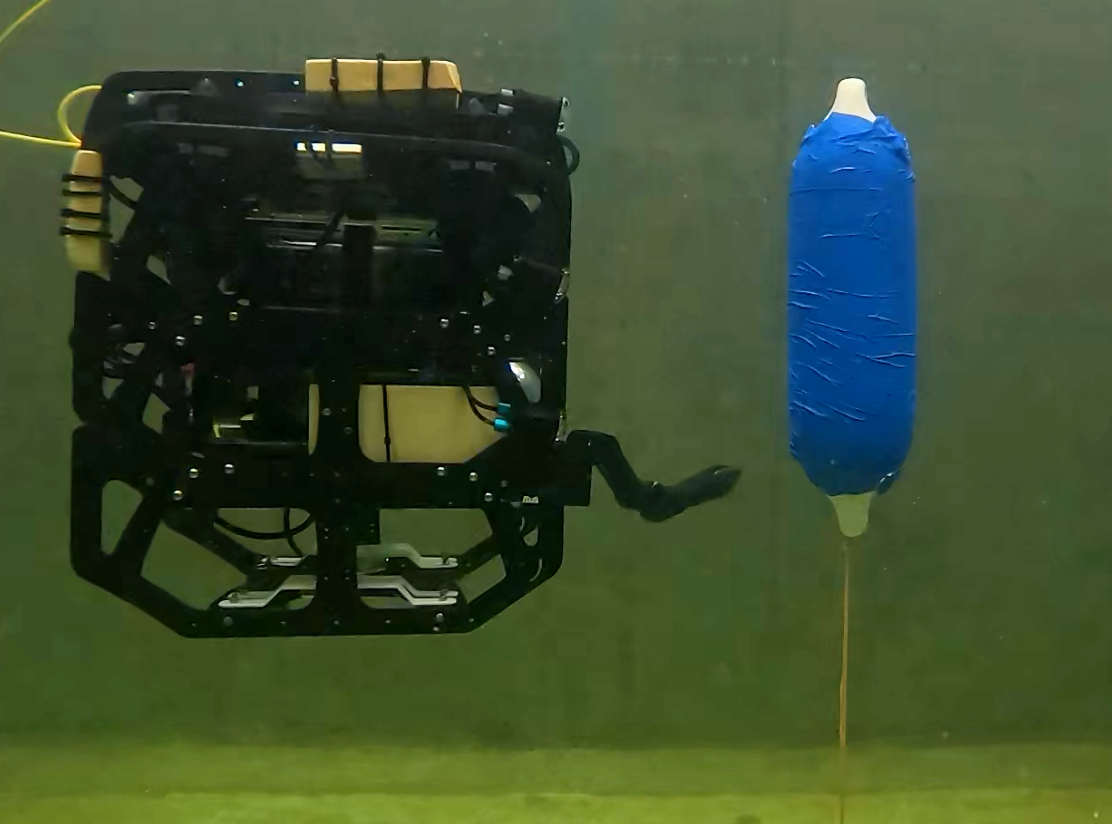}
        \captionsetup{justification=centering}
        \caption{Initial Position}
        \label{fig:inspection_object_rgb}
    \end{subfigure}
    \hfill
    \begin{subfigure}[t]{0.18\textwidth}
        \centering
        \includegraphics[width=\textwidth]{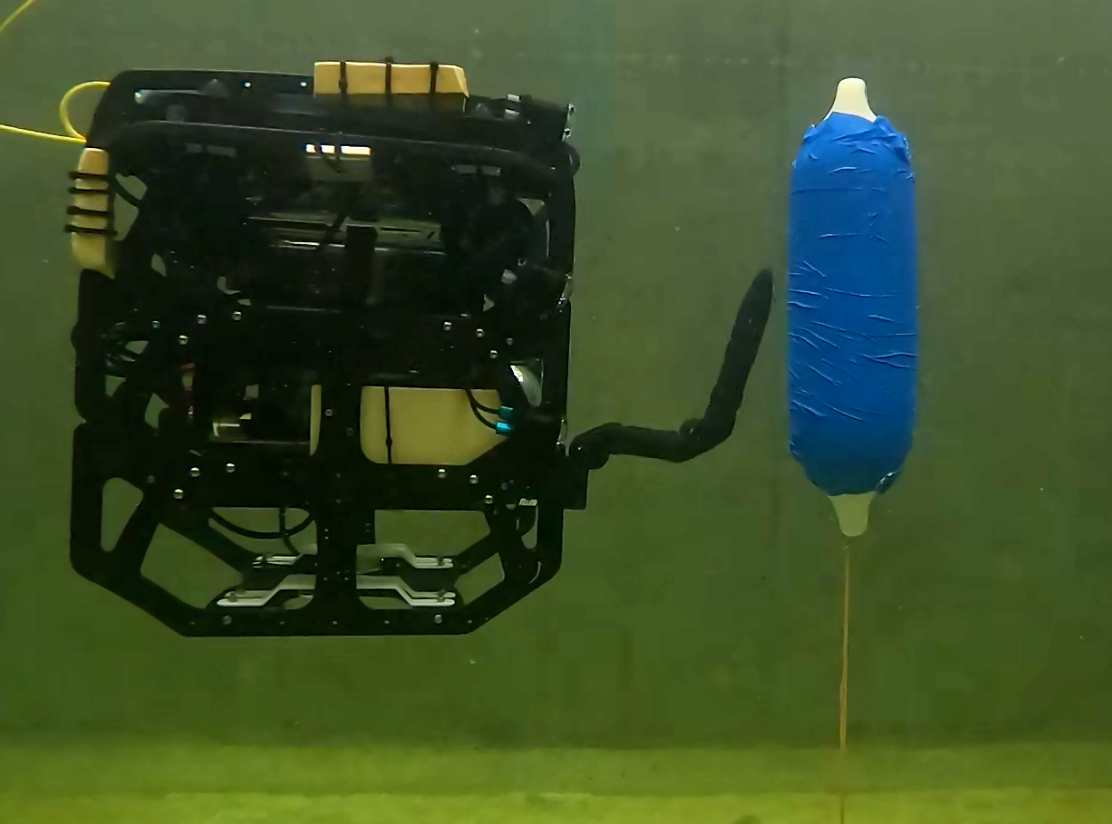}
        \captionsetup{justification=centering}
        \caption{Motion Step 1}
        \label{fig:inspection_object_depth}
    \end{subfigure}
    \hfill
    \begin{subfigure}[t]{0.18\textwidth}
        \centering
        \includegraphics[width=\textwidth]{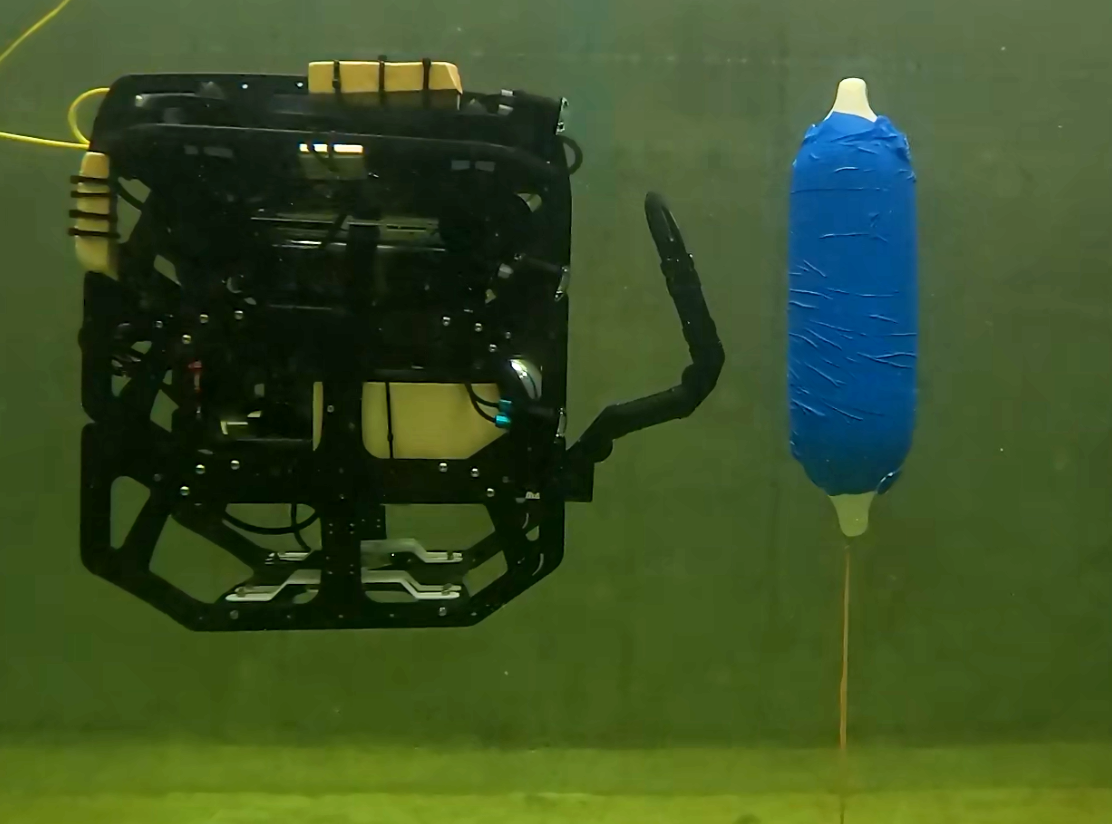}
        \captionsetup{justification=centering}
        \caption{Motion Step 2}
        \label{fig:obstacle_detection_rgb}
    \end{subfigure}
    \hfill
    \begin{subfigure}[t]{0.18\textwidth}
        \centering
        \includegraphics[width=\textwidth]{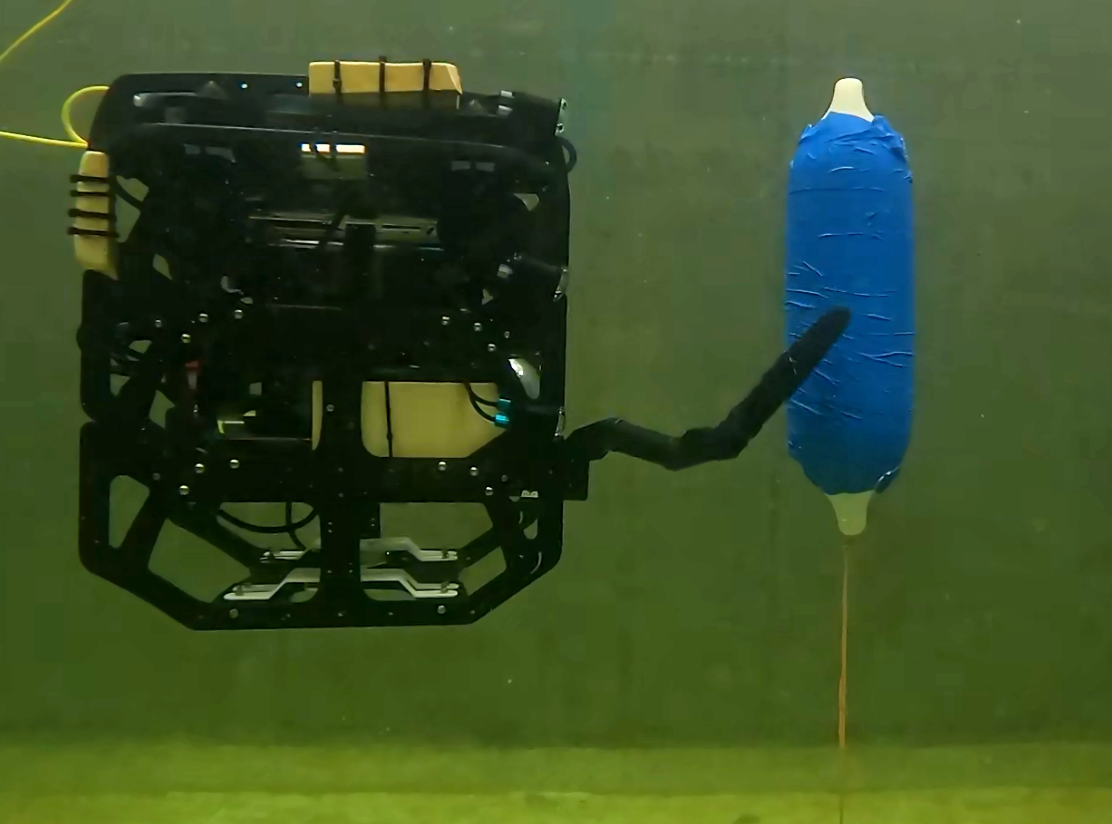}
        \captionsetup{justification=centering}
        \caption{Motion Step 3}
        \label{fig:obstacle_detection_depth}
    \end{subfigure}
    \hfill
    \begin{subfigure}[t]{0.18\textwidth}
        \centering
        \includegraphics[width=\textwidth]{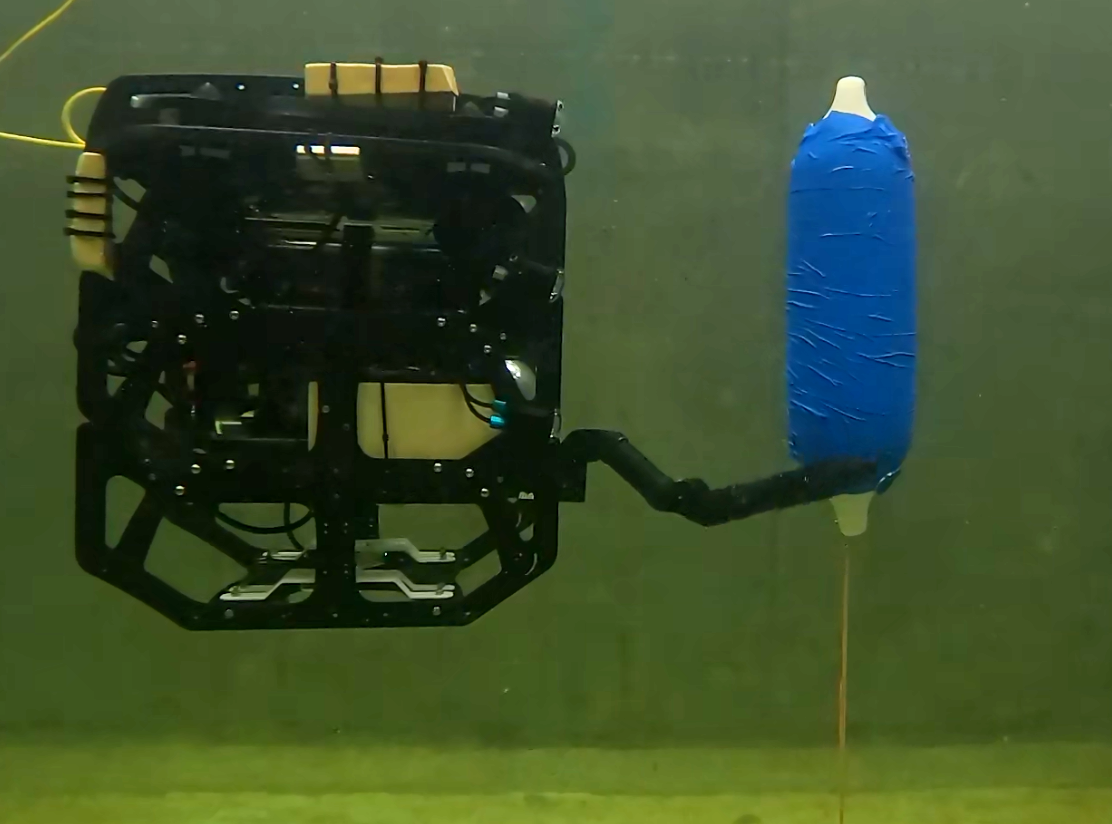}
        \captionsetup{justification=centering}
        \caption{Target Position}
        \label{fig:inspection_process}
    \end{subfigure}
    \caption{
        Snapshots from the experiments illustrate the progression of the UVMS motion.
        \textbf{(a)} initial position,
        \textbf{(b)}, \textbf{(c)}, and \textbf{(d)} show the active obstacle avoidance motion, and
        \textbf{(e)} depicts the UVMS reaching the target position.
    }
    \label{fig:experimental_images}
\end{figure*}

\begin{figure}[t]
    \centering
    \includegraphics[width=\linewidth]{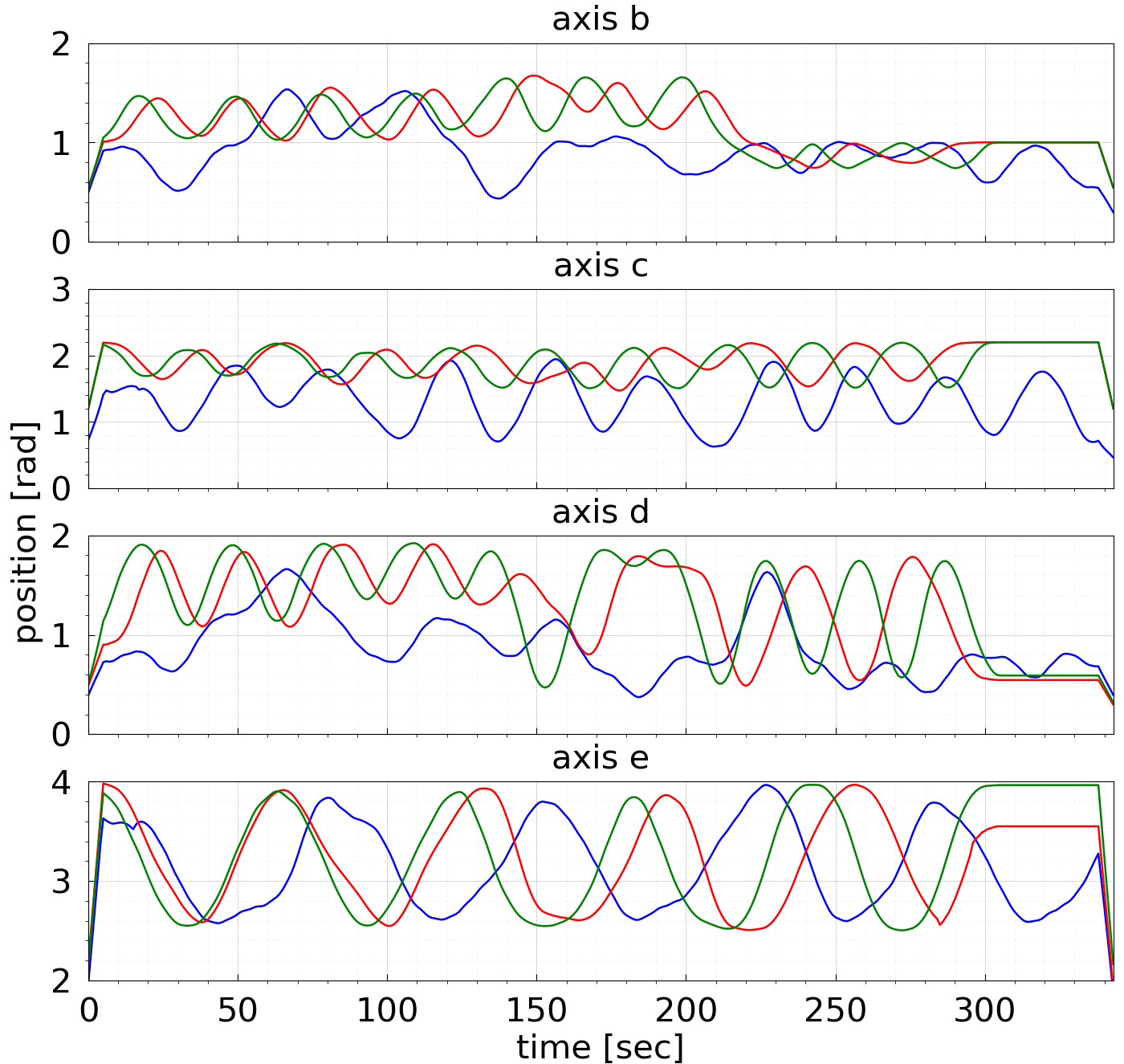}
    \caption{
        Joint position trajectories (10 consecutive goals performed by the UVMS) over time for four joints (b, c, d, e) comparing PRM (green), PRM-HMS (red), and RRT (blue).
    }
    \label{fig:trajectories}
\end{figure}

\begin{table}[t]
    \centering
    \caption{Mean Abs. Error Across Joints (10 Goals)}
    \label{tab:average_errors}
    \begin{tabular}{|c|c|c|c|c|}
        \hline
        \textbf{Test} & \multicolumn{2}{c|}{\textbf{PRM vs. HMS}} & \multicolumn{2}{c|}{\textbf{PRM vs. RRT}} \\
        \cline{2-5}
        & \textbf{Mean [rad]} & \textbf{Std. [rad]} & \textbf{Mean [rad]} & \textbf{Std. [rad]} \\
        \hline
        1 & 0.3522 & 0.3176 & 0.4651 & 0.4552 \\
        \hline
        2 & 0.3658 & 0.2962 & 0.5156 & 0.4629 \\
        \hline
        3 & 0.2757 & 0.4263 & 0.5494 & 0.4738 \\
        \hline
        4 & 0.3287 & 0.2981 & 0.4838 & 0.5627 \\
        \hline
        5 & 0.3867 & 0.4122 & 0.5397 & 0.4837 \\
        \hline
    \end{tabular}
\end{table}

The experimental procedure is divided into two phases: i) HMS training phase, and ii) Goal execution phase. For the HMS Training Phase, we populate the HMS repository by sampling the manipulator’s joint space. A series of virtual goals (i.e., target configurations that are not physically reached) are used to populate the HMS space. This training phase allows the system to learn frequently traversed joint-space segments and to store corresponding motion primitives along with associated metrics such as uncertainty and time estimates.

In the Goal Execution Phase, new target configurations for the manipulator are generated. The algorithm then uses the enhanced PRM, augmented by the learned HMS data, to compute collision-free trajectories toward these goals. An overview of the experimental setup is given in Fig.~\ref{fig:experimental_images}.

The recorded joint trajectories for 10 consecutive goals of the manipulator are shown in Fig.~\ref{fig:trajectories}. This figure compares the trajectories generated by the baseline PRM (green), the RRT algorithm (blue), and our proposed AHMP (red). As observed, the PRM-HMS trajectories closely follow the baseline PRM, demonstrating that our approach effectively replicates the reference motion. In contrast, the RRT trajectories exhibit noticeable deviations, particularly in joints b and d, which is indicative of its distinct exploration strategy.

We conducted five similar motions of these tests, and the average errors, computed as the difference between the recorded and planned trajectories, are summarized in Table~\ref{tab:average_errors}.

The errors between PRM and AHMP are relatively low, confirming the high fidelity of our method in replicating the PRM motion. Conversely, the errors between PRM and RRT are significantly higher, reflecting the fundamental differences in their path planning strategies.

Our findings demonstrate that the AHMP achieves a balance between accuracy and efficiency. The low error rates compared to PRM indicate that the AHMP can effectively replicate high-quality paths, while the observed speed gains (not explicitly shown here but implied by the use of HMS) suggest a significant improvement in computational efficiency. These results validate AHMP as a robust and efficient motion planning solution for underwater robotics.

Across multiple trials with varying start and goal configurations, the system consistently showed low average error ( Table~\ref{tab:average_errors}), demonstrating reliable and repeatable performance in controlled conditions. This paired with the speed gain, makes our proposed methodology extremely useful for motion planning in complex marine environments.

\section{Conclusion}

In this paper, we present an enhanced motion-planning framework for underwater manipulation that combines a HMS with a classical PRM. The HMS caches frequently used ``highway'' paths, significantly reducing search overhead for repeated goals—especially important for tasks like underwater spot welding. A BN refines uncertainty estimates in real-time, guiding the planner to reuse reliable paths under changing conditions. By maintaining a dense PRM, the system retains thorough coverage, yet avoids the computational blow-up of a large search space thanks to HMS shortcuts. Experiments using a BlueROV2 with a Reach Alpha 5 manipulator demonstrate that this approach yields consistent, collision-free trajectories, highlighting its effectiveness for multi-goal tasks in cluttered underwater environments.

Currently, the method has been validated over a set of experiments that mimic applications which require precise arm movements, i.e. for inspection or cleaning tasks that do not require extensive handling of objects. Future works include expanding AHMP to address more complex whole-body manipulation for tasks that involve grasping and manoeuvrability of objects in dynamic underwater environments.

\section*{Acknowledgment}
This work has been supported by the EPSRC project UNderwater IntervenTion for offshore renewable Energies (UNITE) grant number EP/X024806/1

\bibliographystyle{IEEEtran}
\bibliography{reference}

\begin{thebibliography}{10}
\providecommand{\url}[1]{#1}
\csname url@samestyle\endcsname
\providecommand{\newblock}{\relax}
\providecommand{\bibinfo}[2]{#2}
\providecommand{\BIBentrySTDinterwordspacing}{\spaceskip=0pt\relax}
\providecommand{\BIBentryALTinterwordstretchfactor}{4}
\providecommand{\BIBentryALTinterwordspacing}{\spaceskip=\fontdimen2\font plus
\BIBentryALTinterwordstretchfactor\fontdimen3\font minus \fontdimen4\font\relax}
\providecommand{\BIBforeignlanguage}[2]{{%
\expandafter\ifx\csname l@#1\endcsname\relax
\typeout{** WARNING: IEEEtran.bst: No hyphenation pattern has been}%
\typeout{** loaded for the language `#1'. Using the pattern for}%
\typeout{** the default language instead.}%
\else
\language=\csname l@#1\endcsname
\fi
#2}}
\providecommand{\BIBdecl}{\relax}
\BIBdecl

\bibitem{weld_1}
B.~Guo and X.~Li, ``Arc bubble edge detection method based on deep transfer learning in underwater wet welding,'' \emph{Scientific Reports}, vol.~14, no.~1, p. 22628, 2024, published: 2024/09/30.

\bibitem{weld_2}
S.~I. Wahidi, S.~Oterkus, and E.~Oterkus, ``Robotic welding techniques in marine structures and production processes: A systematic literature review,'' \emph{Marine Structures}, vol.~95, p. 103608, 2024.

\bibitem{art_22}
Z.~Zhao, S.~Cheng, Y.~Ding, and Z.~Zhou, ``A survey of optimization-based task and motion planning: From classical to learning approaches,'' \emph{IEEE/ASME Transactions}, 2024.

\bibitem{art_23}
M.~Rastegarpanah, M.~E. Asif, J.~Butt, and H.~Voos, ``Mobile robotics and 3d printing: Addressing challenges in path planning and scalability,'' \emph{Virtual and Physical Prototyping}, 2024.

\bibitem{art_24}
J.~Liu, H.~J. Yap, and A.~S.~M. Khairuddin, ``Review on motion planning of robotic manipulator in dynamic environments,'' \emph{Journal of Sensors}, 2024.

\bibitem{art_1}
Y.~Wang and X.~Guo, ``Memory-based stochastic trajectory optimization for manipulator obstacle avoiding motion planning,'' in \emph{2022 7th Asia-Pacific Conference on Intelligent Robot Systems (ACIRS)}, 2022, pp. 188--194.

\bibitem{art_20}
F.~Nauert and P.~Kampmann, ``Inspection and maintenance of industrial infrastructure with autonomous underwater robots,'' \emph{Frontiers in Robotics and AI}, vol.~10, p. 1240276, 2023.

\bibitem{morgan2022autonomous}
E.~Morgan, I.~Carlucho, W.~Ard, and C.~Barbalata, ``Autonomous underwater manipulation: Current trends in dynamics, control, planning, perception, and future directions,'' \emph{Current Robotics Reports}, vol.~3, no.~4, pp. 187--198, 2022.

\bibitem{art_21}
H.~Tugal, K.~Cetin, X.~Han, I.~Kucukdemiral, J.~Roe, Y.~Petillot, and M.~S. Erden, ``Sliding mode controller for positioning of an underwater vehicle subject to disturbances and time delays,'' in \emph{2022 International Conference on Robotics and Automation (ICRA)}, 2022, pp. 3034--3039.

\bibitem{yoerger1990design}
D.~R. Yoerger and J.~B. Newman, ``Design and control of underwater vehicles: A survey,'' \emph{IFAC proceedings volumes}, vol.~23, no.~12, pp. 113--122, 1990.

\bibitem{whitcomb2000design}
L.~L. Whitcomb, ``Design of auv navigation systems,'' \emph{IEEE journal of oceanic engineering}, vol.~25, no.~2, pp. 177--188, 2000.

\bibitem{allotta2011tethered}
B.~Allotta, A.~Caiti, R.~Costanzi, L.~Fanfani, G.~Monni, L.~Pugi, and A.~Ridolfi, ``Tethered underwater vehicles: Survey and new perspectives,'' \emph{Annual Reviews in Control}, vol.~35, no.~1, pp. 1--22, 2011.

\bibitem{art_9}
G.~Zuo, C.~Wu, and G.~Huang, ``Repetitive path planning with experience-based bidirectional rrt,'' in \emph{Design Advances in Aerospace Robotics}, M.~Ceccarelli, L.~Santo, M.~Paoloni, and G.~Cupertino, Eds.\hskip 1em plus 0.5em minus 0.4em\relax Cham: Springer Nature Switzerland, 2023, pp. 177--192.

\bibitem{art_11}
D.~Honerkamp, T.~Welschehold, and A.~Valada, ``N$^{2}$m$^{2}$: Learning navigation for arbitrary mobile manipulation motions in unseen and dynamic environments,'' \emph{IEEE Transactions on Robotics}, vol.~39, no.~5, p. 3601–3619, Oct. 2023.

\bibitem{art_10}
M.~Moll, C.~Chamzas, Z.~Kingston, and L.~E. Kavraki, ``Hyperplan: A framework for motion planning algorithm selection and parameter optimization,'' in \emph{2021 IEEE/RSJ International Conference on Intelligent Robots and Systems (IROS)}, 2021, pp. 2511--2518.

\bibitem{chomp}
N.~Ratliff, M.~Zucker, J.~A. Bagnell, and S.~Srinivasa, ``Chomp: Gradient optimization techniques for efficient motion planning,'' in \emph{2009 IEEE International Conference on Robotics and Automation}, 2009, pp. 489--494.

\bibitem{sandros}
P.~Chen and Y.~Hwang, ``Sandros: a dynamic graph search algorithm for motion planning,'' \emph{IEEE Transactions on Robotics and Automation}, vol.~14, no.~3, pp. 390--403, 1998.

\bibitem{prm}
L.~Kavraki, P.~Svestka, J.-C. Latombe, and M.~Overmars, ``Probabilistic roadmaps for path planning in high-dimensional configuration spaces,'' \emph{IEEE Transactions on Robotics and Automation}, vol.~12, no.~4, pp. 566--580, 1996.

\bibitem{art_5}
D.~Das, Y.~Lu, E.~Plaku, and X.~Xiao, ``Motion memory: Leveraging past experiences to accelerate future motion planning,'' in \emph{2024 IEEE International Conference on Robotics and Automation (ICRA)}.\hskip 1em plus 0.5em minus 0.4em\relax IEEE, May 2024, p. 16467–16474.

\bibitem{art_27}
I.~Solis, J.~Motes, M.~Qin, M.~Morales, and N.~M. Amato, ``Experience-based subproblem planning for multi-robot motion planning,'' 2024.

\bibitem{art_28}
R.~Terasawa, Y.~Ariki, T.~Narihira, T.~Tsuboi, and K.~Nagasaka, ``3d-cnn based heuristic guided task-space planner for faster motion planning,'' in \emph{2020 IEEE International Conference on Robotics and Automation (ICRA)}, 2020, pp. 9548--9554.

\bibitem{art_6}
X.~Cao, L.~Ren, X.~Wang, and C.~Sun, ``Path re-planning method of unmanned underwater vehicles based on dynamic bayesian threat assessment,'' \emph{Ocean Engineering}, vol. 315, p. 119819, 2025.

\bibitem{art_13}
H.~Xu and J.~Pan, ``Auv motion planning in uncertain flow fields using bayes adaptive mdps,'' \emph{IEEE Robotics and Automation Letters}, vol.~7, no.~2, pp. 5575--5582, 2022.

\bibitem{carlucho2020reinforcement}
I.~Carlucho, M.~De~Paula, C.~Barbalata, and G.~G. Acosta, ``A reinforcement learning control approach for underwater manipulation under position and torque constraints,'' in \emph{Global oceans 2020: Singapore--US gulf coast}.\hskip 1em plus 0.5em minus 0.4em\relax IEEE, 2020, pp. 1--7.

\bibitem{art_4}
D.~Youakim, P.~Cieslak, A.~Dornbush, A.~Palomer, P.~Ridao, and M.~Likhachev, ``Multirepresentation, multiheuristic a* search-based motion planning for a free-floating underwater vehicle-manipulator system in unknown environment,'' \emph{Journal of Field Robotics}, vol.~37, no.~6, pp. 925--950, 2020.

\bibitem{art_18}
B.~Adabala and Z.~Ajanović, ``A multi-heuristic search-based motion planning for automated parking,'' in \emph{2023 XXIX International Conference on Information, Communication and Automation Technologies (ICAT)}.\hskip 1em plus 0.5em minus 0.4em\relax IEEE, Jun. 2023, p. 1–8.

\bibitem{art_17}
J.~Wang, R.~Miao, J.~Ma, and H.~Shi, ``Search-based motion planning with task-space motion primitives for hyper-redundant manipulators in cluttered environment,'' in \emph{2024 6th International Conference on Reconfigurable Mechanisms and Robots (ReMAR)}, 2024, pp. 554--559.

\bibitem{art_12}
R.~Bernardo, J.~M. Sousa, and P.~J. Gonçalves, ``A novel framework to improve motion planning of robotic systems through semantic knowledge-based reasoning,'' \emph{Computers \& Industrial Engineering}, vol. 182, p. 109345, 2023.

\bibitem{bluerobotics}
{Blue Robotics}, ``Blue robotics -- high-quality marine robotics components,'' \url{https://bluerobotics.com/}, 2024.

\bibitem{reach}
{Reach Robotics}, ``Reach robotics -- underwater manipulators,'' \url{https://reachrobotics.com/}, 2024.

\bibitem{Ceres}
\BIBentryALTinterwordspacing
S.~Agarwal, K.~Mierle, and T.~C.~S. Team, ``{Ceres Solver},'' 10 2023. [Online]. Available: \url{https://github.com/ceres-solver/ceres-solver}
\BIBentrySTDinterwordspacing

\end{thebibliography}
\end{document}